\documentclass[manuscript]{acmart}

\newcommand{\calS}{\mathcal{S}}
\newcommand{\calA}{\mathcal{A}}

\newcommand{\calT}{\mathcal{T}}
\newcommand{\calD}{\mathcal{D}}

\newcommand{\mE}{\mathbb{E}}

\usepackage{bbding}
\definecolor{dkred}{rgb}{0.8,0,0}
\definecolor{dkgreen}{rgb}{0,0.4,0}
\definecolor{tickgreen}{rgb}{0,0.6,0}
\newcommand{\tick}{\textcolor{tickgreen}{\CheckmarkBold}}
\newcommand{\fail}{\textcolor{red}{\XSolidBrush}}
\usepackage{multirow}

\AtBeginDocument{%
  \providecommand\BibTeX{{%
    \normalfont B\kern-0.5em{\scshape i\kern-0.25em b}\kern-0.8em\TeX}}}

\setcopyright{acmcopyright}
\copyrightyear{2021}
\acmYear{2021}
\acmDOI{xxx}

\acmConference[SimuRec '21]{RecSys 2021 Workshop on Simulation Methods for Recommender Systems}{October 2, 2021}{Amsterdam}
\acmBooktitle{RecSys 2021 Workshop on Simulation Methods for Recommender Systems, October 2, 2021, Amsterdam}
\acmPrice{xxx}
\acmISBN{xxx}

\begin{document}
\fancyhead{}
\title{
Accelerating Offline Reinforcement Learning Application \\ in Real-Time Bidding and Recommendation: Potential Use of Simulation 
% in Recommender Systems
}


\author{Haruka Kiyohara}
\authornote{This work was done during HK's internship at negocia, Inc.}
\authornote{Corresponding: kiyohara.h.aa@m.titech.ac.jp}
\affiliation{
\institution{Tokyo Institute of Technology}
\city{Tokyo}
\country{Japan}
}
\email{kiyohara.h.aa@m.titech.ac.jp}

\author{Kosuke Kawakami}
% \authornote{kosuke\_kawakami@negocia.jp}
\affiliation{
\institution{negocia, Inc.}
\city{Tokyo}
\country{Japan}
}
\affiliation{
\institution{Tokyo Institute of Technology}
\city{Tokyo}
\country{Japan}
}
\email{kosuke\_kawakami@negocia.jp}

\author{Yuta Saito}
% \authornote{saito@hanjuku-kaso.com}
\affiliation{
\institution{Hanjuku-Kaso Co., Ltd.}
\city{Tokyo}
\country{Japan}
}
\affiliation{
\institution{Cornell University}
\city{Ithaca}
\state{NY}
\country{USA}
}
\email{saito@hanjuku-kaso.com}

% <concept_id>10002951.10003227.10003251</concept_id>
% <concept_desc>Information systems~Multimedia information systems</concept_desc>
% <concept_id>10002951.10003227.10003351</concept_id>
% <concept_desc>Information systems~Data mining</concept_desc>
% <concept_id>10002951.10003317</concept_id>
% <concept_desc>Information systems~Information retrieval</concept_desc>
% <concept_id>10002951.10003317.10003359.10003360</concept_id>
% <concept_desc>Information systems~Test collections</concept_desc>

% \ccsdesc[500]{Information systems~Multimedia information systems}
% \ccsdesc[500]{Information systems~Data mining}
% \ccsdesc[500]{Information systems~Information retrieval}
% \ccsdesc[500]{Information systems~Test collections}

% \keywords{Recommender Systems; Automatic Playlist Continuation; Music Recommendation Systems; Challenge; Benchmark; Evaluation}

\renewcommand{\shortauthors}{Kiyohara, et al}

\newcommand{\kawakami}[1]{\textcolor{red}{[KK: {#1}]}}
\newcommand{\kiyohara}[1]{\textcolor{cyan}{[HK: {#1}]}}
\newcommand{\saito}[1]{\textcolor{blue}{[YS: {#1}]}}

% draft
\begin{abstract}
In recommender systems (RecSys) and real-time bidding (RTB) for online advertisements, we often try to optimize \textit{sequential} decision making using bandit and reinforcement learning (RL) techniques. In these applications, \textit{offline reinforcement learning} (offline RL) and \textit{off-policy evaluation} (OPE) are beneficial because they enable safe policy optimization using only logged data without any risky online interaction. 
In this position paper, we explore the potential of using simulation to accelerate practical research of offline RL and OPE, particularly in RecSys and RTB. Specifically, we discuss how simulation can help us conduct empirical research of offline RL and OPE. 
We take a position to argue that \textit{we should effectively use simulations in the empirical research of offline RL and OPE.} 
To refute the counterclaim that experiments using \textit{only} real-world data are preferable, we first point out the underlying risks and reproducibility issue in real-world experiments. Then, we describe how these issues can be addressed by using simulations. Moreover, we show how to incorporate the benefits of both real-world and simulation-based experiments to defend our position.
Finally, we also present an open challenge to further facilitate practical research of offline RL and OPE in RecSys and RTB, with respect to public simulation platforms.
As a possible solution for the issue, we show our ongoing open source project and its potential use case. We believe that building and utilizing simulation-based evaluation platforms for offline RL and OPE will be of great interest and relevance for the RecSys and RTB community.
\end{abstract}


% \keywords{explicit feedback, implicit feedback, recommender systems, inverse propensity score, counterfactual estimation}

\maketitle

\section{Introduction}
In recommender systems (RecSys) and real-time bidding (RTB) for online advertisements, we often use \textit{sequential} decision making algorithms to increase sales or to enhance user satisfaction. 
For this purpose, interactive bandit and reinforcement learning (RL) are considered powerful tools. 
The RecSys/RTB research communities have studied many applications of bandit and RL and demonstrated their effectiveness in a wide variety of settings~\citep{zhao2019deep, zhao2021dear, zhao2018recommendations, zhao2017deep, zhao2018deep_p, ie2019recsim, cai2017real, wu2018budget, zhao2018deep_s, jin2018real, hao2020dynamic, zou2019reinforcement}.
However, deploying RL policies in real-world systems is often difficult due to the need for risky online interactions. 
Specifically, when we use an adaptive policy and \textit{learn} it in the real environment, numerous numbers of \textit{exploration} is needed before acquiring near-optimal decision makings~\citep{levine2020offline}. The non-optimal exploration is harmful because it may damage sales or user satisfaction~\citep{xiao2021general}. Moreover, we often use online A/B testing to \textit{evaluate} how well a policy works in the real environment. However, it involves high stakes because the unseen new policy may perform poorly on the system~\citep{gilotte2018offline}. 
Therefore, online deployment of RL policies is often limited due to risk concerns, despite their potential benefits after the successful deployment.

Emerging paradigms such as \textit{offline reinforcement learning} (offline RL) and \textit{off-policy evaluation} (OPE) try to tackle these issues in a \textit{data-driven} manner~\citep{levine2020offline}. In offline RL and OPE, we aim to \textit{learn} and \textit{evaluate} a new policy using only previously logged data, without any risky online interaction. 
The major benefit of offline RL and OPE is that we can obtain a new policy that is likely to perform well in a completely safe manner, by 1) learning a new policy using only the logged data (offline RL), and 2) estimating the policy performance using the logged data to guarantee the safety in deployment (OPE).
The potential to reduce the risks in deploying RL policies is gaining researchers' interest. There are many works on offline RL~\citep{levine2020offline, fujimoto2019off, kumar2019stabilizing, kumar2020conservative, fu2020d4rl, gulcehre2020rl, agarwal2020optimistic, argenson2020model, yu2020mopo, kidambi2020morel, paine2020hyperparameter, gulcehre2021regularized, yang2021representation, chen2019bail, yu2021combo} and OPE~\citep{beygelzimer2009offset, precup2000eligibility, strehl2010learning, dudik2014doubly, wang2017optimal, su2020doubly, jiang2016doubly, thomas2016data, saito2020open, voloshin2019empirical, fu2021benchmarks, le2019batch}, and also in their applicability in RecSys practice~\citep{gilotte2018offline, gruson2019offline, santana2020mars, xiao2021general, ma2020off, chen2019top, rohde2018recogym, mazoure2021improving, saito2021evaluating}.

\vspace{2mm}
\textbf{Discussion topic.}
In this paper, we discuss how simulation studies can help accelerate offline RL/OPE research, especially in RecSys/RTB. 
In particular, we focus on the roles of simulations in the evaluation of offline RL/OPE because empirical research is essential for researchers to compare offline RL/OPE methods and analyze their failure cases, leading to a new challenging research direction~\citep{fu2021benchmarks, saito2020open, voloshin2019empirical}. Moreover, validating the performance of the offline RL policies and the accuracy of OPE estimators is crucial to ensure their applicability in real-life situations~\citep{fu2021benchmarks}.

\vspace{2mm}
\textbf{Our position.}
We take a position that \textit{\textbf{we should effectively use simulations for the evaluation of offline RL and OPE.}}
Against the position to argue that \textit{only} the real-world data should be used in the experiments, we first show the difficulties of comprehensive and reproducible experiments incurred in real-world experiments. 
Then, we demonstrate the advantages of simulation-based experiments and how both real-world and simulation-based experiments are important from different perspectives.
Finally, by presenting our ongoing open source project and its expected use case, we show how a simulation platform can assist future offline RL/OPE  research in RecSys/RTB.

\section{Preliminaries}
In (general) RL, we have total $T$ timesteps to optimize our decision making (the special $T=1$ case is called the contextual bandit problem). At every timestep $t$, the decision maker first observes state $s_t \in \calS$ and decide which action $a_t \in \calA$ to take according to the policy $\pi(a_t \mid s_t)$. Then, the decision maker receives a reward $r_t \sim P_r(r_t \mid s_t, a_t)$ and observes the state transition $s_{t+1} \sim \calT(s_{t+1} \mid s_t, a_t)$, where $Pr(\cdot)$ and $\calT(\cdot)$ are the unknown probability distributions. 
For example, in a RecSys setting, $(s_t, a_t, r_t)$ can be user features, an item that the system recommends to the user, and the user's click indicator, respectively.
Here, the objective of RL is to obtain a policy that maximizes the following policy performance (i.e., expected total rewards) $V(\pi) := \mE_{\tau_{\pi}} \left[ \sum_{t=1}^{T} \gamma^{t-1} r_t \right]$, where $\gamma \in (0, 1]$ is a discount factor and $\mE_{\tau_{\pi}}[\cdot]$ is the expectation over the trajectory distribution $p_{\pi}(\tau) \sim p(s_1) \prod_{t=1}^T \pi(a_t \mid s_t) Pr(r_t \mid s_t, a_t) \calT(s_{t+1} \mid s_t, a_t)$.

Let us suppose there is a logged dataset $\calD_{\beta}$ collected by a \textit{behavior} policy $\pi_{\beta}$ as follows.
$$
\calD_{\beta} := \{ \{ (s_t^{(i)}, a_t^{(i)}, r_t^{(i)}, s_{t+1}^{(i)}) \}_{t=1}^T \}_{i=1}^n \sim \prod_{i=1}^n \left( p(s_1) \prod_{t=1}^T \pi_{\beta}(a_t \mid s_t) \calT(s_{t+1} \mid s_t, a_t) \right),
$$
where the dataset consists of $n$ trajectories.
In offline RL, we aim to \textit{learn} a new policy $\pi_{\theta}$ that maximizes the policy performance $V(\pi_{\theta})$ using only $\calD_{\beta}$. 
In OPE, the goal is to \textit{evaluate}, or estimate, the policy performance of a new (evaluation) policy $\pi_{\theta}$ using an OPE estimator $\hat{V}$ and $\calD_{\beta}$ as $\hat{V}(\pi_{\theta}; \calD_{\beta}) \approx V(\pi_{\theta})$. 
To succeed in offline RL/OPE, it is essential to address the \textit{distribution shift} between the new policy $\pi_{\theta}$ and the behavior policy $\pi_{\beta}$. 
Therefore, various algorithms and estimators have been proposed for that purpose~\citep{agarwal2020optimistic, levine2020offline, fujimoto2019off, kumar2019stabilizing, kumar2020conservative, beygelzimer2009offset, argenson2020model, yu2020mopo, kidambi2020morel, paine2020hyperparameter, gulcehre2021regularized, yang2021representation, chen2019bail, yu2021combo, precup2000eligibility, strehl2010learning, dudik2014doubly, wang2017optimal, su2020doubly, jiang2016doubly, thomas2016data, le2019batch}.
To evaluate and compare these methods in empirical research, we need to access both the logged dataset $\calD_{\beta}$ and the ground-truth policy performance of evaluation policy $V(\pi_{\theta})$.
In the following sections, we discuss how we can obtain $\calD_{\beta}$ and  $V(\pi_{\theta})$ to conduct experiments of offline RL/OPE using both real-world and simulation-based synthetic data.

\section{Is the use of real-world data sufficient to facilitate offline RL?}
In this section, we discuss the advantages and drawbacks of the counterclaim: \textit{only the real-world data should be used in experiments of offline RL and OPE.}

We can implement a real-world offline RL/OPE experiment by running (at least) two different policies in the real-world environment.
First, behavior policy $\pi_{\beta}$ collects a logged dataset $\calD_{\beta}$. 
Then, for the evaluation of offline RL/OPE, we need to approximate $V(\pi_{\theta})$ based on \textit{on-policy} estimation of the policy performance, i.e., $V_\mathrm{on}(\pi_{\theta}) := n^{-1} \sum_{i=1}^{n} \sum_{t=1}^T \gamma^{t-1} r_t^{(i)} (\approx V(\pi_{\theta}))$, by deploying evaluation policy $\pi_{\theta}$ to an online environment. 
The advantage of real-world experiments compared to simulation is that it is informative in the sense that the experimental results are expected to generalize in real-world applications~\citep{saito2020open}.

However, there are two critical drawbacks in empirical studies using only real-world data.
The first issue is the risky data collection process and resulting limited experimental settings in the comprehensive experiments. 
The real-world experiments always necessitate the high-cost data collection process because the online interactions can be harmful until the performance of data collection policies ($\pi_{\beta}$ and $\pi_{\theta}$) is guaranteed~\citep{gilotte2018offline}. 
Therefore, it is difficult to deploy a variety of policies due to this risk concern, and the available empirical findings in real-world experiments are often limited. 
For example, when evaluating offline RL algorithms, we often want to know how well the algorithms learn from different logged data, such as the one collected by a sub-optimal policy~\citep{fu2020d4rl, gulcehre2020rl}. However, deploying such a sub-optimal behavior policy is often demanding because it may damage sales or user satisfaction~\citep{levine2020offline, qin2021neorl}. 
Moreover, in the evaluation of OPE estimators, researchers are often curious about how the divergence between behavior and evaluation policies affects the accuracy of the performance estimation~\citep{voloshin2019empirical, saito2021evaluating}. Nonetheless, deploying such largely different evaluation policies is challenging, as there is huge uncertainty in their performance~\citep{levine2020offline}.

The second issue is the lack of reproducibility. Due to confidentiality and data collection costs in RecSys/RTB practice, there is only one public real-world dataset for OPE research (Open Bandit Dataset~\citep{saito2020open}). It is also difficult to publicize a real-world dataset for offline RL because the evaluation of a new policy requires access to the environment~\citep{fu2020d4rl, gulcehre2020rl}. Therefore, conducting a reliable and comprehensive experiment is extremely difficult using only real-world data, which we argue is a bottleneck of the current offline RL/OPE research in RecSys/RTB practice.

\section{How can simulations accelerate offline RL research?}
\label{sec:simulation_experiment}
In this section, we describe how simulations can help evaluate offline RL/OPE methods together with real-world data.

An alternative way to conduct experiments is to build a simulation platform and use it as a substitute for the real environment. Specifically, we can first deploy behavior policy $\pi_{\beta}$ to the simulation environment and obtain a synthetic dataset $\calD_{\beta}^{\ast}$. Then, we can calculate the ground-truth policy performance $V^{\ast}(\pi_{\theta})$ or approximate it by \textit{on-policy} estimation $V^{\ast}_{\mathrm{on}}(\pi_{\theta})$ when the ground-truth calculation is computationally intensive.
The important point here is that the whole experimental procedure does not require any risky online interaction in the real environment. 

Since the policy deployment in the simulation platform is always safe, we can gain abundant findings from simulation research~\citep{fujimoto2019benchmarking, voloshin2019empirical, fu2021benchmarks, xi2021interpretable}. 
For example, in the evaluation of offline RL, we can easily deploy a sub-optimal behavior policy, which is often difficult in real-world experiments~\citep{fu2020d4rl, gulcehre2020rl}. Moreover, we can also analyze the learning process of offline RL by deploying a new policy several times in different training checkpoints, which is challenging due to risks and deployment costs in the real-world experiments~\citep{matsushima2021deployment}. In addition, we can test how well an OPE estimator identifies evaluation policies that perform poorly, which is crucial to avoid failures in practical scenarios~\citep{mcinerney2020counterfactual}. 

Furthermore, we can also tackle the reproducibility issue in real-world experiments by publicizing simulation platforms. 
Using an open-access simulation platform, researchers can easily reproduce the experimental results, which leads to a reliable comparison of the existing works~\citep{fu2020d4rl, gulcehre2020rl}. Therefore, simulation-based experiments are beneficial in enabling reproducible comprehensive studies of offline RL/OPE.

Although the simulation-based empirical research overcomes the drawbacks of real-world experiments, it should also be noted that simulation-based experiments have a \textit{simulation gap} issue~\citep{zhao2020sim}.
Specifically, to model the real environment, we need function approximations for the probability distributions (i.e., $Pr(r_t \mid s_t, a_t)$ and $T(s_{t+1} \mid s_t, a_t)$).
Unfortunately, there must be an inevitable modeling bias which may lead to less informative results.

However, since both real-world and simulation-based experiments have different advantages, we can leverage both for different purposes, as shown in Table~\ref{tab:pros_cons}. 
Specifically, we can first conduct simulation-based comprehensive experiments to see how the configuration changes affect the performance of offline RL/OPE methods to discuss both the advantages and limitations of the methods in a reproducible manner. 
We can also verify if offline RL policies and OPE estimators work in a real-life scenario using real-world experiments with limited online interactions.
Here, by performing preliminary experiments on a simulation platform and removing policies that are likely to perform poorly in advance, we can implement real-world experiments in a less risky manner.
Thus, we argue that \textit{we should effectively use simulations in the empirical research of offline RL and OPE.} 

\begin{table}[tb]
    \centering
    \caption{Comparison of the advantages and usage of real-world and simulation-based experiments}
    \begin{tabular}{c|ccc|c}
        \toprule
        experiment & \textbf{reality} & \textbf{safety} & \textbf{reproducibility}& \textbf{usage}
        \\ \midrule
        real-world & \tick & \fail & \fail & performance verification in real-world \\
        simulation-based & \fail & \tick & \tick & comprehensive study \\ \bottomrule
    \end{tabular}
    \label{tab:pros_cons}
\end{table}

\section{Towards practical research of offline RL in RecSys and RTB}
In this section, we discuss how we can further accelerate offline RL/OPE research in RecSys/RTB practice.

The benefits of the simulation-based experiments have indeed pushed forward the offline RL/OPE research. 
Specifically, many research papers~\citep{kumar2020conservative, argenson2020model, yu2020mopo, kidambi2020morel, paine2020hyperparameter, gulcehre2021regularized, yang2021representation, chen2019bail, yu2021combo, fu2021benchmarks} have been published using a variety of simulated control tasks and their standardized synthetic datasets collected by diverse policies~\citep{fu2020d4rl, gulcehre2020rl}. Moreover, the simulation-based benchmark experiments play important roles for researchers to discuss both advantages and limitations of the existing offline RL and OPE methods~\citep{fujimoto2019benchmarking, xi2021interpretable, voloshin2019empirical, fu2021benchmarks}. 

Practical applications, however, are still limited, especially for offline RL (such as~\citep{qin2021neorl, zhan2021deepthermal, xiao2021general, ma2020off, chen2019top, mazoure2021improving}). We attribute this to the lack of application-specific simulation environments that provide useful insights for specific research questions.
For example, RecSys/RTB are unique regarding their huge action space and highly stochastic and delayed rewards~\citep{ma2020off, cai2017real, zou2019reinforcement}. 
Therefore, we need to build a simulation platform imitating such specific characteristics to better understand the empirical performance of offline RL/OPE methods in these particular situations.

In the RecSys setting, there are two dominant simulation platforms well-designed for offline RL/OPE research, OpenBanditPipeline (OBP)~\citep{saito2020open} and RecoGym~\citep{rohde2018recogym}. They are both beneficial in enabling simulation-based experiments in a fully offline manner. Moreover, OBP is helpful in practice because it provides streamlined implementations of the experimental procedure and the modules to preprocess real-world data. However, their limitation is that they are unable to handle RL policies.\footnote{Specifically, OBP can handle only contextual bandits, and RecoGym combines contextual bandits and non-Markov interactions called \textit{organic} session.} Therefore, with these currently available packages, it is difficult to evaluate the offline RL and OPE methods of RL policies relevant to the real-world sequential decision makings~\citep{xiao2021general}. Moreover, there are no such simulation platforms in RTB. There is a need to build a simulation-based evaluation platform for offline RL and OPE in RecSys/RTB settings.

\begin{figure}[tb]
    \centering
    \includegraphics[width=0.75\linewidth]{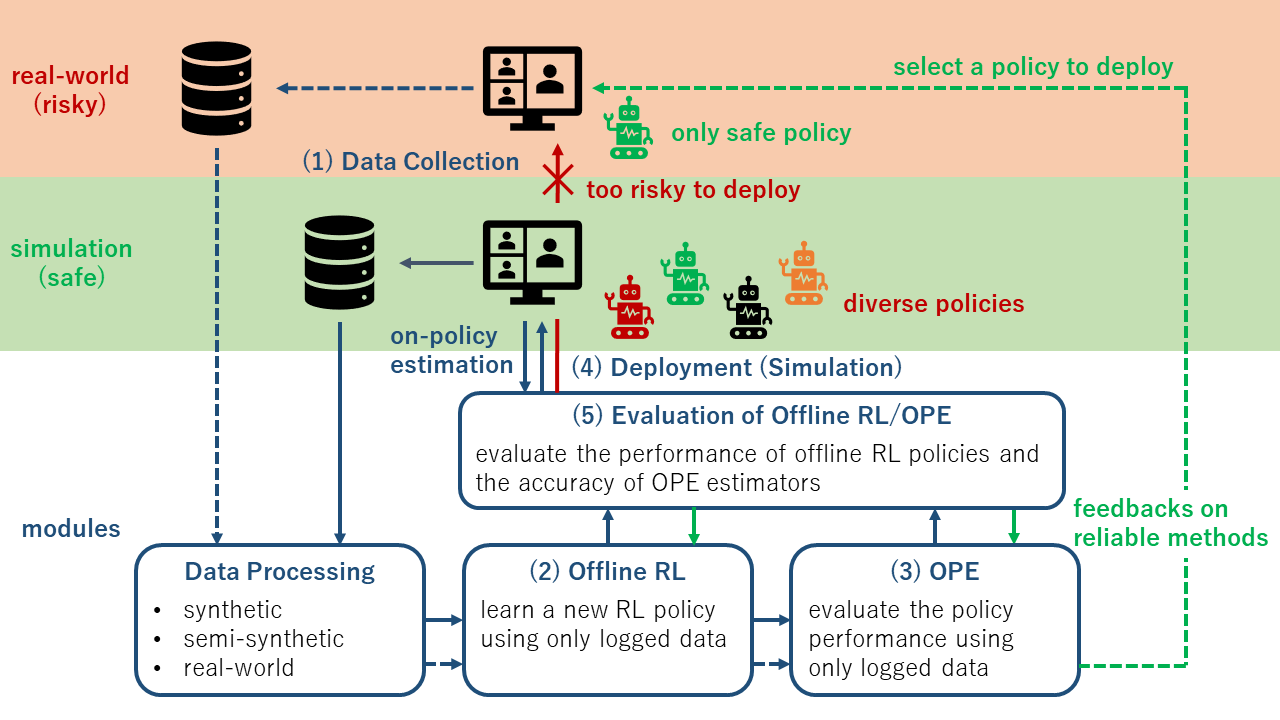}
    \caption{Overview of our simulation platform and its workflow}
    \label{fig:gym}
\end{figure}

Motivated by the above necessity, we are developing an open-source simulation platform in the RTB setting. Our design principle is to provide an easy-to-use platform to the users. 
Below, we present an expected use case and describe how to utilize our platform in offline RL/OPE empirical research.

We aim to conduct both simulation-based and real-world experiments, as described in Section~\ref{sec:simulation_experiment}.
The solid arrows in Figure~\ref{fig:gym} show the workflow of simulation-based comprehensive experiments based on our platform.
The key feature of the platform is that there are design choices for researchers, such as what behavior and evaluation policies to use and what offline RL and OPE methods to test. Moreover, researchers can also customize the environmental configurations in the simulation platform, such as action space $\calA$ and total timestep $T$, to see how the configuration changes affect the performance of offline RL/OPE.
After the detailed investigation in simulation-based experiments, we can also verify if the offline RL/OPE methods work in real-life scenarios with a limited number of online interactions. Our platform also provides streamlined implementation and data processing modules for assisting real-world experiments, as shown with the dotted arrows in Figure~\ref{fig:gym}. The platform also allows researchers to identify a safe policy in advance using our semi-synthetic simulation, which replicates the real environment based on the original real-world dataset. The results of such a semi-synthetic simulation may help reduce the risks in real-world experiments.

Finally, since we plan to publicize the platform, the research community can engage in our project to make the simulation platform to be a more diverse benchmark and more practically relevant.
Moreover, we plan to extend our platform to the RecSys setting. 
These additional efforts will allow researchers to easily involve in the empirical research of offline RL/OPE in RecSys/RTB.

\bibliographystyle{ACM-Reference-Format}
\bibliography{ref.bib}


\begin{thebibliography}{55}

%%% the Web address.

\ifx \showCODEN    \undefined \def \showCODEN     #1{\unskip}     \fi
\ifx \showDOI      \undefined \def \showDOI       #1{#1}\fi
\ifx \showISBNx    \undefined \def \showISBNx     #1{\unskip}     \fi
\ifx \showISBNxiii \undefined \def \showISBNxiii  #1{\unskip}     \fi
\ifx \showISSN     \undefined \def \showISSN      #1{\unskip}     \fi
\ifx \showLCCN     \undefined \def \showLCCN      #1{\unskip}     \fi
\ifx \shownote     \undefined \def \shownote      #1{#1}          \fi
\ifx \showarticletitle \undefined \def \showarticletitle #1{#1}   \fi
\ifx \showURL      \undefined \def \showURL       {\relax}        \fi
% invisible to TeX
\providecommand\bibfield[2]{#2}
\providecommand\bibinfo[2]{#2}
\providecommand\natexlab[1]{#1}
\providecommand\showeprint[2][]{arXiv:#2}

\bibitem[\protect\citeauthoryear{Agarwal, Schuurmans, and Norouzi}{Agarwal
  et~al\mbox{.}}{2020}]%
        {agarwal2020optimistic}
\bibfield{author}{\bibinfo{person}{Rishabh Agarwal}, \bibinfo{person}{Dale
  Schuurmans}, {and} \bibinfo{person}{Mohammad Norouzi}.}
  \bibinfo{year}{2020}\natexlab{}.
\newblock \showarticletitle{An optimistic perspective on offline reinforcement
  learning}. In \bibinfo{booktitle}{\emph{Proceedings of the 37th International
  Conference on Machine Learning}}, Vol.~\bibinfo{volume}{119}.
  \bibinfo{pages}{104--114}.
\newblock


\bibitem[\protect\citeauthoryear{Argenson and Dulac-Arnold}{Argenson and
  Dulac-Arnold}{2021}]%
        {argenson2020model}
\bibfield{author}{\bibinfo{person}{Arthur Argenson} {and}
  \bibinfo{person}{Gabriel Dulac-Arnold}.} \bibinfo{year}{2021}\natexlab{}.
\newblock \showarticletitle{Model-based offline planning}. In
  \bibinfo{booktitle}{\emph{International Conference on Learning
  Representations}}.
\newblock


\bibitem[\protect\citeauthoryear{Beygelzimer and Langford}{Beygelzimer and
  Langford}{2009}]%
        {beygelzimer2009offset}
\bibfield{author}{\bibinfo{person}{Alina Beygelzimer} {and}
  \bibinfo{person}{John Langford}.} \bibinfo{year}{2009}\natexlab{}.
\newblock \showarticletitle{The offset tree for learning with partial labels}.
  In \bibinfo{booktitle}{\emph{Proceedings of the 15th ACM SIGKDD International
  Conference on Knowledge Discovery and Data Mining}}.
  \bibinfo{pages}{129--138}.
\newblock


\bibitem[\protect\citeauthoryear{Cai, Ren, Zhang, Malialis, Wang, Yu, and
  Guo}{Cai et~al\mbox{.}}{2017}]%
        {cai2017real}
\bibfield{author}{\bibinfo{person}{Han Cai}, \bibinfo{person}{Kan Ren},
  \bibinfo{person}{Weinan Zhang}, \bibinfo{person}{Kleanthis Malialis},
  \bibinfo{person}{Jun Wang}, \bibinfo{person}{Yong Yu}, {and}
  \bibinfo{person}{Defeng Guo}.} \bibinfo{year}{2017}\natexlab{}.
\newblock \showarticletitle{Real-time bidding by reinforcement learning in
  display advertising}. In \bibinfo{booktitle}{\emph{Proceedings of the 10th
  ACM International Conference on Web Search and Data Mining}}.
  \bibinfo{pages}{661--670}.
\newblock


\bibitem[\protect\citeauthoryear{Chen, Beutel, Covington, Jain, Belletti, and
  Chi}{Chen et~al\mbox{.}}{2019a}]%
        {chen2019top}
\bibfield{author}{\bibinfo{person}{Minmin Chen}, \bibinfo{person}{Alex Beutel},
  \bibinfo{person}{Paul Covington}, \bibinfo{person}{Sagar Jain},
  \bibinfo{person}{Francois Belletti}, {and} \bibinfo{person}{Ed~H Chi}.}
  \bibinfo{year}{2019}\natexlab{a}.
\newblock \showarticletitle{Top-k off-policy correction for a REINFORCE
  recommender system}. In \bibinfo{booktitle}{\emph{Proceedings of the 12th ACM
  International Conference on Web Search and Data Mining}}.
  \bibinfo{pages}{456--464}.
\newblock


\bibitem[\protect\citeauthoryear{Chen, Zhou, Wang, Wang, Wu, and Ross}{Chen
  et~al\mbox{.}}{2019b}]%
        {chen2019bail}
\bibfield{author}{\bibinfo{person}{Xinyue Chen}, \bibinfo{person}{Zijian Zhou},
  \bibinfo{person}{Zheng Wang}, \bibinfo{person}{Che Wang},
  \bibinfo{person}{Yanqiu Wu}, {and} \bibinfo{person}{Keith Ross}.}
  \bibinfo{year}{2019}\natexlab{b}.
\newblock \showarticletitle{BAIL: Best-action imitation learning for batch deep
  reinforcement learning}. In \bibinfo{booktitle}{\emph{Advances in Neural
  Information Processing Systems}}, Vol.~\bibinfo{volume}{33}.
  \bibinfo{pages}{18353--18363}.
\newblock


\bibitem[\protect\citeauthoryear{Dud{\'\i}k, Erhan, Langford, and
  Li}{Dud{\'\i}k et~al\mbox{.}}{2014}]%
        {dudik2014doubly}
\bibfield{author}{\bibinfo{person}{Miroslav Dud{\'\i}k},
  \bibinfo{person}{Dumitru Erhan}, \bibinfo{person}{John Langford}, {and}
  \bibinfo{person}{Lihong Li}.} \bibinfo{year}{2014}\natexlab{}.
\newblock \showarticletitle{Doubly robust policy evaluation and optimization}.
\newblock \bibinfo{journal}{\emph{Statist. Sci.}} \bibinfo{volume}{29},
  \bibinfo{number}{4} (\bibinfo{year}{2014}), \bibinfo{pages}{485--511}.
\newblock


\bibitem[\protect\citeauthoryear{Fu, Kumar, Nachum, Tucker, and Levine}{Fu
  et~al\mbox{.}}{2020}]%
        {fu2020d4rl}
\bibfield{author}{\bibinfo{person}{Justin Fu}, \bibinfo{person}{Aviral Kumar},
  \bibinfo{person}{Ofir Nachum}, \bibinfo{person}{George Tucker}, {and}
  \bibinfo{person}{Sergey Levine}.} \bibinfo{year}{2020}\natexlab{}.
\newblock \showarticletitle{D4RL: Datasets for deep data-driven reinforcement
  learning}.
\newblock \bibinfo{journal}{\emph{arXiv preprint arXiv:2004.07219}}
  (\bibinfo{year}{2020}).
\newblock


\bibitem[\protect\citeauthoryear{Fu, Norouzi, Nachum, Tucker, Wang, Novikov,
  Yang, Zhang, Chen, Kumar, et~al\mbox{.}}{Fu et~al\mbox{.}}{2021}]%
        {fu2021benchmarks}
\bibfield{author}{\bibinfo{person}{Justin Fu}, \bibinfo{person}{Mohammad
  Norouzi}, \bibinfo{person}{Ofir Nachum}, \bibinfo{person}{George Tucker},
  \bibinfo{person}{Ziyu Wang}, \bibinfo{person}{Alexander Novikov},
  \bibinfo{person}{Mengjiao Yang}, \bibinfo{person}{Michael~R Zhang},
  \bibinfo{person}{Yutian Chen}, \bibinfo{person}{Aviral Kumar},
  {et~al\mbox{.}}} \bibinfo{year}{2021}\natexlab{}.
\newblock \showarticletitle{Benchmarks for deep off-policy evaluation}.
\newblock \bibinfo{journal}{\emph{arXiv preprint arXiv:2103.16596}}
  (\bibinfo{year}{2021}).
\newblock


\bibitem[\protect\citeauthoryear{Fujimoto, Conti, Ghavamzadeh, and
  Pineau}{Fujimoto et~al\mbox{.}}{2019a}]%
        {fujimoto2019benchmarking}
\bibfield{author}{\bibinfo{person}{Scott Fujimoto}, \bibinfo{person}{Edoardo
  Conti}, \bibinfo{person}{Mohammad Ghavamzadeh}, {and} \bibinfo{person}{Joelle
  Pineau}.} \bibinfo{year}{2019}\natexlab{a}.
\newblock \showarticletitle{Benchmarking batch deep reinforcement learning
  algorithms}.
\newblock \bibinfo{journal}{\emph{arXiv preprint arXiv:1910.01708}}
  (\bibinfo{year}{2019}).
\newblock


\bibitem[\protect\citeauthoryear{Fujimoto, Meger, and Precup}{Fujimoto
  et~al\mbox{.}}{2019b}]%
        {fujimoto2019off}
\bibfield{author}{\bibinfo{person}{Scott Fujimoto}, \bibinfo{person}{David
  Meger}, {and} \bibinfo{person}{Doina Precup}.}
  \bibinfo{year}{2019}\natexlab{b}.
\newblock \showarticletitle{Off-policy deep reinforcement learning without
  exploration}. In \bibinfo{booktitle}{\emph{Proceedings of the 36th
  International Conference on Machine Learning}}, Vol.~\bibinfo{volume}{97}.
  \bibinfo{pages}{2052--2062}.
\newblock


\bibitem[\protect\citeauthoryear{Gilotte, Calauz{\`e}nes, Nedelec, Abraham, and
  Doll{\'e}}{Gilotte et~al\mbox{.}}{2018}]%
        {gilotte2018offline}
\bibfield{author}{\bibinfo{person}{Alexandre Gilotte},
  \bibinfo{person}{Cl{\'e}ment Calauz{\`e}nes}, \bibinfo{person}{Thomas
  Nedelec}, \bibinfo{person}{Alexandre Abraham}, {and} \bibinfo{person}{Simon
  Doll{\'e}}.} \bibinfo{year}{2018}\natexlab{}.
\newblock \showarticletitle{Offline a/b testing for recommender systems}. In
  \bibinfo{booktitle}{\emph{Proceedings of the 11th ACM International
  Conference on Web Search and Data Mining}}. \bibinfo{pages}{198--206}.
\newblock


\bibitem[\protect\citeauthoryear{Gruson, Chandar, Charbuillet, McInerney,
  Hansen, Tardieu, and Carterette}{Gruson et~al\mbox{.}}{2019}]%
        {gruson2019offline}
\bibfield{author}{\bibinfo{person}{Alois Gruson}, \bibinfo{person}{Praveen
  Chandar}, \bibinfo{person}{Christophe Charbuillet}, \bibinfo{person}{James
  McInerney}, \bibinfo{person}{Samantha Hansen}, \bibinfo{person}{Damien
  Tardieu}, {and} \bibinfo{person}{Ben Carterette}.}
  \bibinfo{year}{2019}\natexlab{}.
\newblock \showarticletitle{Offline evaluation to make decisions about playlist
  recommendation algorithms}. In \bibinfo{booktitle}{\emph{Proceedings of the
  12th ACM International Conference on Web Search and Data Mining}}.
  \bibinfo{pages}{420--428}.
\newblock


\bibitem[\protect\citeauthoryear{Gulcehre, Colmenarejo, Wang, Sygnowski, Paine,
  Zolna, Chen, Hoffman, Pascanu, and de~Freitas}{Gulcehre
  et~al\mbox{.}}{2021}]%
        {gulcehre2021regularized}
\bibfield{author}{\bibinfo{person}{Caglar Gulcehre},
  \bibinfo{person}{Sergio~G{\'o}mez Colmenarejo}, \bibinfo{person}{Ziyu Wang},
  \bibinfo{person}{Jakub Sygnowski}, \bibinfo{person}{Thomas Paine},
  \bibinfo{person}{Konrad Zolna}, \bibinfo{person}{Yutian Chen},
  \bibinfo{person}{Matthew Hoffman}, \bibinfo{person}{Razvan Pascanu}, {and}
  \bibinfo{person}{Nando de Freitas}.} \bibinfo{year}{2021}\natexlab{}.
\newblock \showarticletitle{Regularized behavior value estimation}.
\newblock \bibinfo{journal}{\emph{arXiv preprint arXiv:2103.09575}}
  (\bibinfo{year}{2021}).
\newblock


\bibitem[\protect\citeauthoryear{Gulcehre, Wang, Novikov, Paine, G{\'o}mez,
  Zolna, Agarwal, Merel, Mankowitz, Paduraru, et~al\mbox{.}}{Gulcehre
  et~al\mbox{.}}{2020}]%
        {gulcehre2020rl}
\bibfield{author}{\bibinfo{person}{Caglar Gulcehre}, \bibinfo{person}{Ziyu
  Wang}, \bibinfo{person}{Alexander Novikov}, \bibinfo{person}{Thomas Paine},
  \bibinfo{person}{Sergio G{\'o}mez}, \bibinfo{person}{Konrad Zolna},
  \bibinfo{person}{Rishabh Agarwal}, \bibinfo{person}{Josh~S Merel},
  \bibinfo{person}{Daniel~J Mankowitz}, \bibinfo{person}{Cosmin Paduraru},
  {et~al\mbox{.}}} \bibinfo{year}{2020}\natexlab{}.
\newblock \showarticletitle{RL Unplugged: A Collection of Benchmarks for
  Offline Reinforcement Learning}.
\newblock \bibinfo{journal}{\emph{Advances in Neural Information Processing
  Systems}}  \bibinfo{volume}{33}, \bibinfo{pages}{7248--7259}.
\newblock


\bibitem[\protect\citeauthoryear{Hao, Peng, Ma, Wang, Jin, Hao, Chen, Bai, Xie,
  Xu, Zheng, Yu, Li, Xu, and Gai}{Hao et~al\mbox{.}}{2020}]%
        {hao2020dynamic}
\bibfield{author}{\bibinfo{person}{Xiaotian Hao}, \bibinfo{person}{Zhaoqing
  Peng}, \bibinfo{person}{Yi Ma}, \bibinfo{person}{Guan Wang},
  \bibinfo{person}{Junqi Jin}, \bibinfo{person}{Jianye Hao},
  \bibinfo{person}{Shan Chen}, \bibinfo{person}{Rongquan Bai},
  \bibinfo{person}{Mingzhou Xie}, \bibinfo{person}{Miao Xu},
  \bibinfo{person}{Zhenzhe Zheng}, \bibinfo{person}{Chuan Yu},
  \bibinfo{person}{Han Li}, \bibinfo{person}{Jian Xu}, {and}
  \bibinfo{person}{Kun Gai}.} \bibinfo{year}{2020}\natexlab{}.
\newblock \showarticletitle{Dynamic knapsack optimization towards efficient
  multi-channel sequential advertising}. In
  \bibinfo{booktitle}{\emph{Proceedings of the 37th International Conference on
  Machine Learning}}. \bibinfo{pages}{4060--4070}.
\newblock


\bibitem[\protect\citeauthoryear{Ie, Hsu, Mladenov, Jain, Narvekar, Wang, Wu,
  and Boutilier}{Ie et~al\mbox{.}}{2019}]%
        {ie2019recsim}
\bibfield{author}{\bibinfo{person}{Eugene Ie}, \bibinfo{person}{Chih-wei Hsu},
  \bibinfo{person}{Martin Mladenov}, \bibinfo{person}{Vihan Jain},
  \bibinfo{person}{Sanmit Narvekar}, \bibinfo{person}{Jing Wang},
  \bibinfo{person}{Rui Wu}, {and} \bibinfo{person}{Craig Boutilier}.}
  \bibinfo{year}{2019}\natexlab{}.
\newblock \showarticletitle{RecSim: A configurable simulation platform for
  recommender systems}.
\newblock \bibinfo{journal}{\emph{arXiv preprint arXiv:1909.04847}}
  (\bibinfo{year}{2019}).
\newblock


\bibitem[\protect\citeauthoryear{Jiang and Li}{Jiang and Li}{2016}]%
        {jiang2016doubly}
\bibfield{author}{\bibinfo{person}{Nan Jiang} {and} \bibinfo{person}{Lihong
  Li}.} \bibinfo{year}{2016}\natexlab{}.
\newblock \showarticletitle{Doubly robust off-policy value evaluation for
  reinforcement learning}. In \bibinfo{booktitle}{\emph{Proceedings of the 33rd
  International Conference on International Conference on Machine Learning}},
  Vol.~\bibinfo{volume}{48}. \bibinfo{pages}{652--661}.
\newblock


\bibitem[\protect\citeauthoryear{Jin, Song, Li, Gai, Wang, and Zhang}{Jin
  et~al\mbox{.}}{2018}]%
        {jin2018real}
\bibfield{author}{\bibinfo{person}{Junqi Jin}, \bibinfo{person}{Chengru Song},
  \bibinfo{person}{Han Li}, \bibinfo{person}{Kun Gai}, \bibinfo{person}{Jun
  Wang}, {and} \bibinfo{person}{Weinan Zhang}.}
  \bibinfo{year}{2018}\natexlab{}.
\newblock \showarticletitle{Real-time bidding with multi-agent reinforcement
  learning in display advertising}. In \bibinfo{booktitle}{\emph{Proceedings of
  the 27th ACM International Conference on Information and Knowledge
  Management}}. \bibinfo{pages}{2193--2201}.
\newblock


\bibitem[\protect\citeauthoryear{Kidambi, Rajeswaran, Netrapalli, and
  Joachims}{Kidambi et~al\mbox{.}}{2020}]%
        {kidambi2020morel}
\bibfield{author}{\bibinfo{person}{Rahul Kidambi}, \bibinfo{person}{Aravind
  Rajeswaran}, \bibinfo{person}{Praneeth Netrapalli}, {and}
  \bibinfo{person}{Thorsten Joachims}.} \bibinfo{year}{2020}\natexlab{}.
\newblock \showarticletitle{MOReL: Model-based offline reinforcement learning}.
  In \bibinfo{booktitle}{\emph{Advances in Neural Information Processing
  Systems}}, Vol.~\bibinfo{volume}{33}. \bibinfo{pages}{21810--21823}.
\newblock


\bibitem[\protect\citeauthoryear{Kumar, Fu, Soh, Tucker, and Levine}{Kumar
  et~al\mbox{.}}{2019}]%
        {kumar2019stabilizing}
\bibfield{author}{\bibinfo{person}{Aviral Kumar}, \bibinfo{person}{Justin Fu},
  \bibinfo{person}{Matthew Soh}, \bibinfo{person}{George Tucker}, {and}
  \bibinfo{person}{Sergey Levine}.} \bibinfo{year}{2019}\natexlab{}.
\newblock \showarticletitle{Stabilizing off-policy q-learning via bootstrapping
  error reduction}. In \bibinfo{booktitle}{\emph{Advances in Neural Information
  Processing Systems}}, Vol.~\bibinfo{volume}{32}.
  \bibinfo{pages}{11784--11794}.
\newblock


\bibitem[\protect\citeauthoryear{Kumar, Zhou, Tucker, and Levine}{Kumar
  et~al\mbox{.}}{2020}]%
        {kumar2020conservative}
\bibfield{author}{\bibinfo{person}{Aviral Kumar}, \bibinfo{person}{Aurick
  Zhou}, \bibinfo{person}{George Tucker}, {and} \bibinfo{person}{Sergey
  Levine}.} \bibinfo{year}{2020}\natexlab{}.
\newblock \showarticletitle{Conservative q-learning for offline reinforcement
  learning}. In \bibinfo{booktitle}{\emph{Advances in Neural Information
  Processing Systems}}, Vol.~\bibinfo{volume}{33}. \bibinfo{pages}{1179--1191}.
\newblock


\bibitem[\protect\citeauthoryear{Le, Voloshin, and Yue}{Le
  et~al\mbox{.}}{2019}]%
        {le2019batch}
\bibfield{author}{\bibinfo{person}{Hoang Le}, \bibinfo{person}{Cameron
  Voloshin}, {and} \bibinfo{person}{Yisong Yue}.}
  \bibinfo{year}{2019}\natexlab{}.
\newblock \showarticletitle{Batch policy learning under constraints}. In
  \bibinfo{booktitle}{\emph{Proceedings of the 36th International Conference on
  Machine Learning}}, Vol.~\bibinfo{volume}{97}. \bibinfo{pages}{3703--3712}.
\newblock


\bibitem[\protect\citeauthoryear{{Levine}, {Kumar}, {Tucker}, and
  {Fu}}{{Levine} et~al\mbox{.}}{2020}]%
        {levine2020offline}
\bibfield{author}{\bibinfo{person}{Sergey {Levine}}, \bibinfo{person}{Aviral
  {Kumar}}, \bibinfo{person}{George {Tucker}}, {and} \bibinfo{person}{Justin
  {Fu}}.} \bibinfo{year}{2020}\natexlab{}.
\newblock \showarticletitle{Offline reinforcement learning: tutorial, review,
  and perspectives on open problems}.
\newblock \bibinfo{journal}{\emph{arXiv preprint arXiv:2005.01643}}
  (\bibinfo{year}{2020}).
\newblock


\bibitem[\protect\citeauthoryear{Ma, Zhao, Yi, Yang, Chen, Tang, Hong, and
  Chi}{Ma et~al\mbox{.}}{2020}]%
        {ma2020off}
\bibfield{author}{\bibinfo{person}{Jiaqi Ma}, \bibinfo{person}{Zhe Zhao},
  \bibinfo{person}{Xinyang Yi}, \bibinfo{person}{Ji Yang},
  \bibinfo{person}{Minmin Chen}, \bibinfo{person}{Jiaxi Tang},
  \bibinfo{person}{Lichan Hong}, {and} \bibinfo{person}{Ed~H Chi}.}
  \bibinfo{year}{2020}\natexlab{}.
\newblock \showarticletitle{Off-policy learning in two-stage recommender
  systems}. In \bibinfo{booktitle}{\emph{Proceedings of The Web Conference
  2020}}. \bibinfo{pages}{463--473}.
\newblock


\bibitem[\protect\citeauthoryear{Matsushima, Furuta, Matsuo, Nachum, and
  Gu}{Matsushima et~al\mbox{.}}{2021}]%
        {matsushima2021deployment}
\bibfield{author}{\bibinfo{person}{Tatsuya Matsushima}, \bibinfo{person}{Hiroki
  Furuta}, \bibinfo{person}{Yutaka Matsuo}, \bibinfo{person}{Ofir Nachum},
  {and} \bibinfo{person}{Shixiang Gu}.} \bibinfo{year}{2021}\natexlab{}.
\newblock \showarticletitle{Deployment-efficient reinforcement learning via
  model-based offline optimization}. In \bibinfo{booktitle}{\emph{International
  Conference on Learning Representations}}.
\newblock


\bibitem[\protect\citeauthoryear{Mazoure, Mineiro, Srinath, Sedeh, Precup, and
  Swaminathan}{Mazoure et~al\mbox{.}}{2021}]%
        {mazoure2021improving}
\bibfield{author}{\bibinfo{person}{Bogdan Mazoure}, \bibinfo{person}{Paul
  Mineiro}, \bibinfo{person}{Pavithra Srinath}, \bibinfo{person}{Reza~Sharifi
  Sedeh}, \bibinfo{person}{Doina Precup}, {and} \bibinfo{person}{Adith
  Swaminathan}.} \bibinfo{year}{2021}\natexlab{}.
\newblock \showarticletitle{Improving long-term metrics in recommendation
  systems using short-horizon offline RL}.
\newblock \bibinfo{journal}{\emph{arXiv preprint arXiv:2106.00589}}
  (\bibinfo{year}{2021}).
\newblock


\bibitem[\protect\citeauthoryear{McInerney, Brost, Chandar, Mehrotra, and
  Carterette}{McInerney et~al\mbox{.}}{2020}]%
        {mcinerney2020counterfactual}
\bibfield{author}{\bibinfo{person}{James McInerney}, \bibinfo{person}{Brian
  Brost}, \bibinfo{person}{Praveen Chandar}, \bibinfo{person}{Rishabh
  Mehrotra}, {and} \bibinfo{person}{Benjamin Carterette}.}
  \bibinfo{year}{2020}\natexlab{}.
\newblock \showarticletitle{Counterfactual Evaluation of Slate Recommendations
  with Sequential Reward Interactions}. In
  \bibinfo{booktitle}{\emph{Proceedings of the 26th ACM SIGKDD International
  Conference on Knowledge Discovery and Data Mining}}.
  \bibinfo{pages}{1779--1788}.
\newblock


\bibitem[\protect\citeauthoryear{Paine, Paduraru, Michi, Gulcehre, Zolna,
  Novikov, Wang, and de~Freitas}{Paine et~al\mbox{.}}{2020}]%
        {paine2020hyperparameter}
\bibfield{author}{\bibinfo{person}{Tom~Le Paine}, \bibinfo{person}{Cosmin
  Paduraru}, \bibinfo{person}{Andrea Michi}, \bibinfo{person}{Caglar Gulcehre},
  \bibinfo{person}{Konrad Zolna}, \bibinfo{person}{Alexander Novikov},
  \bibinfo{person}{Ziyu Wang}, {and} \bibinfo{person}{Nando de Freitas}.}
  \bibinfo{year}{2020}\natexlab{}.
\newblock \showarticletitle{Hyperparameter selection for offline reinforcement
  learning}.
\newblock \bibinfo{journal}{\emph{arXiv preprint arXiv:2007.09055}}
  (\bibinfo{year}{2020}).
\newblock


\bibitem[\protect\citeauthoryear{Precup, Sutton, and Singh}{Precup
  et~al\mbox{.}}{2000}]%
        {precup2000eligibility}
\bibfield{author}{\bibinfo{person}{Doina Precup}, \bibinfo{person}{Richard~S.
  Sutton}, {and} \bibinfo{person}{Satinder Singh}.}
  \bibinfo{year}{2000}\natexlab{}.
\newblock \showarticletitle{Eligibility traces for off-policy policy
  evaluation}. In \bibinfo{booktitle}{\emph{Proceedings of the 17th
  International Conference on Machine Learning}}. \bibinfo{pages}{759--766}.
\newblock


\bibitem[\protect\citeauthoryear{Qin, Gao, Zhang, Xu, Huang, Li, Zhang, and
  Yu}{Qin et~al\mbox{.}}{2021}]%
        {qin2021neorl}
\bibfield{author}{\bibinfo{person}{Rongjun Qin}, \bibinfo{person}{Songyi Gao},
  \bibinfo{person}{Xingyuan Zhang}, \bibinfo{person}{Zhen Xu},
  \bibinfo{person}{Shengkai Huang}, \bibinfo{person}{Zewen Li},
  \bibinfo{person}{Weinan Zhang}, {and} \bibinfo{person}{Yang Yu}.}
  \bibinfo{year}{2021}\natexlab{}.
\newblock \showarticletitle{NeoRL: A near real-world benchmark for offline
  reinforcement learning}.
\newblock \bibinfo{journal}{\emph{arXiv preprint arXiv:2102.00714}}
  (\bibinfo{year}{2021}).
\newblock


\bibitem[\protect\citeauthoryear{Rohde, Bonner, Dunlop, Vasile, and
  Karatzoglou}{Rohde et~al\mbox{.}}{2018}]%
        {rohde2018recogym}
\bibfield{author}{\bibinfo{person}{David Rohde}, \bibinfo{person}{Stephen
  Bonner}, \bibinfo{person}{Travis Dunlop}, \bibinfo{person}{Flavian Vasile},
  {and} \bibinfo{person}{Alexandros Karatzoglou}.}
  \bibinfo{year}{2018}\natexlab{}.
\newblock \showarticletitle{RecoGym: A reinforcement learning environment for
  the problem of product recommendation in online advertising}.
\newblock \bibinfo{journal}{\emph{arXiv preprint arXiv:1808.00720}}
  (\bibinfo{year}{2018}).
\newblock


\bibitem[\protect\citeauthoryear{Saito, Aihara, Matsutani, and Narita}{Saito
  et~al\mbox{.}}{2020}]%
        {saito2020open}
\bibfield{author}{\bibinfo{person}{Yuta Saito}, \bibinfo{person}{Shunsuke
  Aihara}, \bibinfo{person}{Megumi Matsutani}, {and} \bibinfo{person}{Yusuke
  Narita}.} \bibinfo{year}{2020}\natexlab{}.
\newblock \showarticletitle{Open Bandit Dataset and Pipeline: Towards realistic
  and reproducible off-policy evaluation}.
\newblock \bibinfo{journal}{\emph{arXiv preprint arXiv:2008.07146}}
  (\bibinfo{year}{2020}).
\newblock


\bibitem[\protect\citeauthoryear{Saito, Udagawa, Kiyohara, Mogi, Narita, and
  Tateno}{Saito et~al\mbox{.}}{2021}]%
        {saito2021evaluating}
\bibfield{author}{\bibinfo{person}{Yuta Saito}, \bibinfo{person}{Takuma
  Udagawa}, \bibinfo{person}{Haruka Kiyohara}, \bibinfo{person}{Kazuki Mogi},
  \bibinfo{person}{Yusuke Narita}, {and} \bibinfo{person}{Kei Tateno}.}
  \bibinfo{year}{2021}\natexlab{}.
\newblock \showarticletitle{Evaluating the Robustness of Off-Policy
  Evaluation}.
\newblock \bibinfo{journal}{\emph{arXiv preprint arXiv:2108.13703}}
  (\bibinfo{year}{2021}).
\newblock


\bibitem[\protect\citeauthoryear{Santana, Melo, Camargo, Brand{\~a}o, Soares,
  Oliveira, and Caetano}{Santana et~al\mbox{.}}{2020}]%
        {santana2020mars}
\bibfield{author}{\bibinfo{person}{Marlesson~RO Santana},
  \bibinfo{person}{Luckeciano~C Melo}, \bibinfo{person}{Fernando~HF Camargo},
  \bibinfo{person}{Bruno Brand{\~a}o}, \bibinfo{person}{Anderson Soares},
  \bibinfo{person}{Renan~M Oliveira}, {and} \bibinfo{person}{Sandor Caetano}.}
  \bibinfo{year}{2020}\natexlab{}.
\newblock \showarticletitle{MARS-Gym: A gym framework to model, train, and
  evaluate recommender systems for marketplaces}. In
  \bibinfo{booktitle}{\emph{2020 International Conference on Data Mining
  Workshops (ICDMW)}}. \bibinfo{pages}{189--197}.
\newblock


\bibitem[\protect\citeauthoryear{Strehl, Langford, Li, and Kakade}{Strehl
  et~al\mbox{.}}{2010}]%
        {strehl2010learning}
\bibfield{author}{\bibinfo{person}{Alex Strehl}, \bibinfo{person}{John
  Langford}, \bibinfo{person}{Lihong Li}, {and} \bibinfo{person}{Sham~M
  Kakade}.} \bibinfo{year}{2010}\natexlab{}.
\newblock \showarticletitle{Learning from logged implicit exploration data}. In
  \bibinfo{booktitle}{\emph{Advances in Neural Information Processing
  Systems}}, Vol.~\bibinfo{volume}{23}. \bibinfo{pages}{2217--2225}.
\newblock


\bibitem[\protect\citeauthoryear{Su, Dimakopoulou, Krishnamurthy, and
  Dud{\'\i}k}{Su et~al\mbox{.}}{2020}]%
        {su2020doubly}
\bibfield{author}{\bibinfo{person}{Yi Su}, \bibinfo{person}{Maria
  Dimakopoulou}, \bibinfo{person}{Akshay Krishnamurthy}, {and}
  \bibinfo{person}{Miroslav Dud{\'\i}k}.} \bibinfo{year}{2020}\natexlab{}.
\newblock \showarticletitle{Doubly robust off-policy evaluation with
  shrinkage}. In \bibinfo{booktitle}{\emph{Proceedings of the 37th
  International Conference on Machine Learning}}, Vol.~\bibinfo{volume}{119}.
  \bibinfo{pages}{9167--9176}.
\newblock


\bibitem[\protect\citeauthoryear{Thomas and Brunskill}{Thomas and
  Brunskill}{2016}]%
        {thomas2016data}
\bibfield{author}{\bibinfo{person}{Philip Thomas} {and} \bibinfo{person}{Emma
  Brunskill}.} \bibinfo{year}{2016}\natexlab{}.
\newblock \showarticletitle{Data-efficient off-policy policy evaluation for
  reinforcement learning}. In \bibinfo{booktitle}{\emph{Proceedings of the 33rd
  International Conference on Machine Learning}}. \bibinfo{pages}{2139--2148}.
\newblock


\bibitem[\protect\citeauthoryear{Voloshin, Le, Jiang, and Yue}{Voloshin
  et~al\mbox{.}}{2019}]%
        {voloshin2019empirical}
\bibfield{author}{\bibinfo{person}{Cameron Voloshin}, \bibinfo{person}{Hoang~M
  Le}, \bibinfo{person}{Nan Jiang}, {and} \bibinfo{person}{Yisong Yue}.}
  \bibinfo{year}{2019}\natexlab{}.
\newblock \showarticletitle{Empirical study of off-policy policy evaluation for
  reinforcement learning}.
\newblock \bibinfo{journal}{\emph{arXiv preprint arXiv:1911.06854}}
  (\bibinfo{year}{2019}).
\newblock


\bibitem[\protect\citeauthoryear{Wang, Agarwal, and Dud{\i}k}{Wang
  et~al\mbox{.}}{2017}]%
        {wang2017optimal}
\bibfield{author}{\bibinfo{person}{Yu-Xiang Wang}, \bibinfo{person}{Alekh
  Agarwal}, {and} \bibinfo{person}{Miroslav Dud{\i}k}.}
  \bibinfo{year}{2017}\natexlab{}.
\newblock \showarticletitle{Optimal and adaptive off-policy evaluation in
  contextual bandits}. In \bibinfo{booktitle}{\emph{Proceedings of the 34th
  International Conference on Machine Learning}}, Vol.~\bibinfo{volume}{70}.
  \bibinfo{pages}{3589--3597}.
\newblock


\bibitem[\protect\citeauthoryear{Wu, Chen, Yang, Wang, Tan, Zhang, Xu, and
  Gai}{Wu et~al\mbox{.}}{2018}]%
        {wu2018budget}
\bibfield{author}{\bibinfo{person}{Di Wu}, \bibinfo{person}{Xiujun Chen},
  \bibinfo{person}{Xun Yang}, \bibinfo{person}{Hao Wang}, \bibinfo{person}{Qing
  Tan}, \bibinfo{person}{Xiaoxun Zhang}, \bibinfo{person}{Jian Xu}, {and}
  \bibinfo{person}{Kun Gai}.} \bibinfo{year}{2018}\natexlab{}.
\newblock \showarticletitle{Budget constrained bidding by model-free
  reinforcement learning in display advertising}. In
  \bibinfo{booktitle}{\emph{Proceedings of the 27th ACM International
  Conference on Information and Knowledge Management}}.
  \bibinfo{pages}{1443--1451}.
\newblock


\bibitem[\protect\citeauthoryear{Xi, Tang, Shen, Liu, Xiong, and Li}{Xi
  et~al\mbox{.}}{2021}]%
        {xi2021interpretable}
\bibfield{author}{\bibinfo{person}{Chenyang Xi}, \bibinfo{person}{Bo Tang},
  \bibinfo{person}{Jiajun Shen}, \bibinfo{person}{Xinfu Liu},
  \bibinfo{person}{Feiyu Xiong}, {and} \bibinfo{person}{Xueying Li}.}
  \bibinfo{year}{2021}\natexlab{}.
\newblock \showarticletitle{Interpretable performance analysis towards offline
  reinforcement learning: A dataset perspective}.
\newblock \bibinfo{journal}{\emph{arXiv preprint arXiv:2105.05473}}
  (\bibinfo{year}{2021}).
\newblock


\bibitem[\protect\citeauthoryear{Xiao and Wang}{Xiao and Wang}{2021}]%
        {xiao2021general}
\bibfield{author}{\bibinfo{person}{Teng Xiao} {and} \bibinfo{person}{Donglin
  Wang}.} \bibinfo{year}{2021}\natexlab{}.
\newblock \showarticletitle{A general offline reinforcement learning framework
  for interactive recommendation}. In \bibinfo{booktitle}{\emph{Proceedings of
  the AAAI Conference on Artificial Intelligence}}, Vol.~\bibinfo{volume}{35}.
  \bibinfo{pages}{4512--4520}.
\newblock


\bibitem[\protect\citeauthoryear{Yang and Nachum}{Yang and Nachum}{2021}]%
        {yang2021representation}
\bibfield{author}{\bibinfo{person}{Mengjiao Yang} {and} \bibinfo{person}{Ofir
  Nachum}.} \bibinfo{year}{2021}\natexlab{}.
\newblock \showarticletitle{Representation matters: Offline pretraining for
  sequential decision making}.
\newblock \bibinfo{journal}{\emph{arXiv preprint arXiv:2102.05815}}
  (\bibinfo{year}{2021}).
\newblock


\bibitem[\protect\citeauthoryear{Yu, Kumar, Rafailov, Rajeswaran, Levine, and
  Finn}{Yu et~al\mbox{.}}{2021}]%
        {yu2021combo}
\bibfield{author}{\bibinfo{person}{Tianhe Yu}, \bibinfo{person}{Aviral Kumar},
  \bibinfo{person}{Rafael Rafailov}, \bibinfo{person}{Aravind Rajeswaran},
  \bibinfo{person}{Sergey Levine}, {and} \bibinfo{person}{Chelsea Finn}.}
  \bibinfo{year}{2021}\natexlab{}.
\newblock \showarticletitle{COMBO: Conservative offline model-based policy
  optimization}.
\newblock \bibinfo{journal}{\emph{arXiv preprint arXiv:2102.08363}}
  (\bibinfo{year}{2021}).
\newblock


\bibitem[\protect\citeauthoryear{Yu, Thomas, Yu, Ermon, Zou, Levine, Finn, and
  Ma}{Yu et~al\mbox{.}}{2020}]%
        {yu2020mopo}
\bibfield{author}{\bibinfo{person}{Tianhe Yu}, \bibinfo{person}{Garrett
  Thomas}, \bibinfo{person}{Lantao Yu}, \bibinfo{person}{Stefano Ermon},
  \bibinfo{person}{James Zou}, \bibinfo{person}{Sergey Levine},
  \bibinfo{person}{Chelsea Finn}, {and} \bibinfo{person}{Tengyu Ma}.}
  \bibinfo{year}{2020}\natexlab{}.
\newblock \showarticletitle{MOPO: Model-based offline policy optimization}. In
  \bibinfo{booktitle}{\emph{Advances in Neural Information Processing
  Systems}}, Vol.~\bibinfo{volume}{33}. \bibinfo{pages}{14129--14142}.
\newblock


\bibitem[\protect\citeauthoryear{Zhan, Xu, Zhang, Huo, Zhu, Yin, and
  Zheng}{Zhan et~al\mbox{.}}{2021}]%
        {zhan2021deepthermal}
\bibfield{author}{\bibinfo{person}{Xianyuan Zhan}, \bibinfo{person}{Haoran Xu},
  \bibinfo{person}{Yue Zhang}, \bibinfo{person}{Yusen Huo},
  \bibinfo{person}{Xiangyu Zhu}, \bibinfo{person}{Honglei Yin}, {and}
  \bibinfo{person}{Yu Zheng}.} \bibinfo{year}{2021}\natexlab{}.
\newblock \showarticletitle{DeepThermal: Combustion optimization for thermal
  power generating units using offline reinforcement learning}.
\newblock \bibinfo{journal}{\emph{arXiv preprint arXiv:2102.11492}}
  (\bibinfo{year}{2021}).
\newblock


\bibitem[\protect\citeauthoryear{Zhao, Qiu, Guan, Zhao, and He}{Zhao
  et~al\mbox{.}}{2018a}]%
        {zhao2018deep_s}
\bibfield{author}{\bibinfo{person}{Jun Zhao}, \bibinfo{person}{Guang Qiu},
  \bibinfo{person}{Ziyu Guan}, \bibinfo{person}{Wei Zhao}, {and}
  \bibinfo{person}{Xiaofei He}.} \bibinfo{year}{2018}\natexlab{a}.
\newblock \showarticletitle{Deep reinforcement learning for sponsored search
  real-time bidding}. In \bibinfo{booktitle}{\emph{Proceedings of the 24th ACM
  SIGKDD International Conference on Knowledge Discovery and Data mining}}.
  \bibinfo{pages}{1021--1030}.
\newblock


\bibitem[\protect\citeauthoryear{Zhao, Queralta, and Westerlund}{Zhao
  et~al\mbox{.}}{2020}]%
        {zhao2020sim}
\bibfield{author}{\bibinfo{person}{Wenshuai Zhao},
  \bibinfo{person}{Jorge~Pe{\~n}a Queralta}, {and} \bibinfo{person}{Tomi
  Westerlund}.} \bibinfo{year}{2020}\natexlab{}.
\newblock \showarticletitle{Sim-to-real transfer in deep reinforcement learning
  for robotics: a survey}. In \bibinfo{booktitle}{\emph{2020 IEEE Symposium
  Series on Computational Intelligence (SSCI)}}. \bibinfo{pages}{737--744}.
\newblock


\bibitem[\protect\citeauthoryear{Zhao, Gu, Zhang, Yang, Liu, Liu, and
  Tang}{Zhao et~al\mbox{.}}{2021}]%
        {zhao2021dear}
\bibfield{author}{\bibinfo{person}{Xiangyu Zhao}, \bibinfo{person}{Changsheng
  Gu}, \bibinfo{person}{Haoshenglun Zhang}, \bibinfo{person}{Xiwang Yang},
  \bibinfo{person}{Xiaobing Liu}, \bibinfo{person}{Hui Liu}, {and}
  \bibinfo{person}{Jiliang Tang}.} \bibinfo{year}{2021}\natexlab{}.
\newblock \showarticletitle{DEAR: Deep reinforcement learning for online
  advertising impression in recommender systems}. In
  \bibinfo{booktitle}{\emph{Proceedings of the AAAI Conference on Artificial
  Intelligence}}, Vol.~\bibinfo{volume}{35}. \bibinfo{pages}{750--758}.
\newblock


\bibitem[\protect\citeauthoryear{Zhao, Xia, Tang, and Yin}{Zhao
  et~al\mbox{.}}{2019}]%
        {zhao2019deep}
\bibfield{author}{\bibinfo{person}{Xiangyu Zhao}, \bibinfo{person}{Long Xia},
  \bibinfo{person}{Jiliang Tang}, {and} \bibinfo{person}{Dawei Yin}.}
  \bibinfo{year}{2019}\natexlab{}.
\newblock \showarticletitle{Deep reinforcement learning for search,
  recommendation, and online advertising: a survey}.
\newblock \bibinfo{journal}{\emph{ACM SIGWEB Newsletter}}
  \bibinfo{number}{Spring} (\bibinfo{year}{2019}), \bibinfo{pages}{1--15}.
\newblock


\bibitem[\protect\citeauthoryear{Zhao, Xia, Zhang, Ding, Yin, and Tang}{Zhao
  et~al\mbox{.}}{2018b}]%
        {zhao2018deep_p}
\bibfield{author}{\bibinfo{person}{Xiangyu Zhao}, \bibinfo{person}{Long Xia},
  \bibinfo{person}{Liang Zhang}, \bibinfo{person}{Zhuoye Ding},
  \bibinfo{person}{Dawei Yin}, {and} \bibinfo{person}{Jiliang Tang}.}
  \bibinfo{year}{2018}\natexlab{b}.
\newblock \showarticletitle{Deep reinforcement learning for page-wise
  recommendations}. In \bibinfo{booktitle}{\emph{Proceedings of the 12th ACM
  Conference on Recommender Systems}}. \bibinfo{pages}{95--103}.
\newblock


\bibitem[\protect\citeauthoryear{Zhao, Zhang, Ding, Xia, Tang, and Yin}{Zhao
  et~al\mbox{.}}{2018c}]%
        {zhao2018recommendations}
\bibfield{author}{\bibinfo{person}{Xiangyu Zhao}, \bibinfo{person}{Liang
  Zhang}, \bibinfo{person}{Zhuoye Ding}, \bibinfo{person}{Long Xia},
  \bibinfo{person}{Jiliang Tang}, {and} \bibinfo{person}{Dawei Yin}.}
  \bibinfo{year}{2018}\natexlab{c}.
\newblock \showarticletitle{Recommendations with negative feedback via pairwise
  deep reinforcement learning}. In \bibinfo{booktitle}{\emph{Proceedings of the
  24th ACM SIGKDD International Conference on Knowledge Discovery and Data
  Mining}}. \bibinfo{pages}{1040--1048}.
\newblock


\bibitem[\protect\citeauthoryear{Zhao, Zhang, Xia, Ding, Yin, and Tang}{Zhao
  et~al\mbox{.}}{2017}]%
        {zhao2017deep}
\bibfield{author}{\bibinfo{person}{Xiangyu Zhao}, \bibinfo{person}{Liang
  Zhang}, \bibinfo{person}{Long Xia}, \bibinfo{person}{Zhuoye Ding},
  \bibinfo{person}{Dawei Yin}, {and} \bibinfo{person}{Jiliang Tang}.}
  \bibinfo{year}{2017}\natexlab{}.
\newblock \showarticletitle{Deep reinforcement learning for list-wise
  recommendations}.
\newblock \bibinfo{journal}{\emph{arXiv preprint arXiv:1801.00209}}
  (\bibinfo{year}{2017}).
\newblock


\bibitem[\protect\citeauthoryear{Zou, Xia, Ding, Song, Liu, and Yin}{Zou
  et~al\mbox{.}}{2019}]%
        {zou2019reinforcement}
\bibfield{author}{\bibinfo{person}{Lixin Zou}, \bibinfo{person}{Long Xia},
  \bibinfo{person}{Zhuoye Ding}, \bibinfo{person}{Jiaxing Song},
  \bibinfo{person}{Weidong Liu}, {and} \bibinfo{person}{Dawei Yin}.}
  \bibinfo{year}{2019}\natexlab{}.
\newblock \showarticletitle{Reinforcement learning to optimize long-term user
  engagement in recommender systems}. In \bibinfo{booktitle}{\emph{Proceedings
  of the 25th ACM SIGKDD International Conference on Knowledge Discovery and
  Data Mining}}. \bibinfo{pages}{2810--2818}.
\newblock


\end{thebibliography}

% \clearpage
% \appendix

\end{document}